\documentclass[sigconf,screen]{acmart}
\usepackage{booktabs}
\usepackage{multirow}
\usepackage[normalem]{ulem}
\useunder{\uline}{\ul}{}
\usepackage{makecell}
\usepackage{pifont}
\usepackage{bbding}
\usepackage{enumitem}
\AtBeginDocument{%
  }

\setcopyright{acmlicensed}
\copyrightyear{2025}
\acmYear{2025}
\acmDOI{XXXXXXX.XXXXXXX}
\acmConference[Conference acronym 'ACMMM]{Make sure to enter the correct
  conference title from your rights confirmation email}{June 03--05,
  2025}{Woodstock, NY}
\acmISBN{978-1-4503-XXXX-X/2018/06}

\begin{document}

\title[ComplexBench-Edit]{ComplexBench-Edit: Benchmarking Complex Instruction-Driven Image Editing via Compositional Dependencies}

\author{Chenglin Wang$^1$, ~Yucheng Zhou$^2$, ~Qianning Wang$^3$, ~Zhe Wang$^1$, ~Kai Zhang$^1$} 
\affiliation{
  \institution{$^1$East China Normal University \country{China}, 
~$^2$University of Macau \country{China}, \\
$^3$Auckland University of Technology \country{New Zealand}}
}
\email{52275901013@stu.ecnu.edu.cn, yucheng.zhou@connect.um.edu.mo}

\renewcommand{\shortauthors}{Wang and Zhou et al.}

\begin{abstract}
Text-driven image editing has achieved remarkable success in following single instructions. However, real-world scenarios often involve complex, multi-step instructions, particularly ``chain'' instructions where operations are interdependent. Current models struggle with these intricate directives, and existing benchmarks inadequately evaluate such capabilities. Specifically, they often overlook multi-instruction and chain-instruction complexities, and common consistency metrics are flawed. To address this, we introduce \textbf{ComplexBench-Edit}, a novel benchmark designed to systematically assess model performance on complex, multi-instruction, and chain-dependent image editing tasks. ComplexBench-Edit also features a new vision consistency evaluation method that accurately assesses non-modified regions by excluding edited areas. Furthermore, we propose a simple yet powerful Chain-of-Thought (CoT)-based approach that significantly enhances the ability of existing models to follow complex instructions. Our extensive experiments demonstrate ComplexBench-Edit's efficacy in differentiating model capabilities and highlight the superior performance of our CoT-based method in handling complex edits. The data and code are released at \url{https://github.com/llllly26/ComplexBench-Edit}.
\end{abstract}






\maketitle

\section{Introduction}
Text-driven diffusion models have led to remarkable success in image generation and editing~\cite{esser2024scaling, brooks2023instructpix2pix, rombach2022high}. These models have demonstrated impressive capabilities in following single, explicit instructions to modify images, achieving high-fidelity and semantically coherent results~\cite{brooks2023instructpix2pix, zhang2023magicbrush, sheynin2024emu}. This progress has significantly empowered users to manipulate visual content with natural language.

\begin{figure}[t]
    \centering
    \includegraphics[width=1\linewidth]{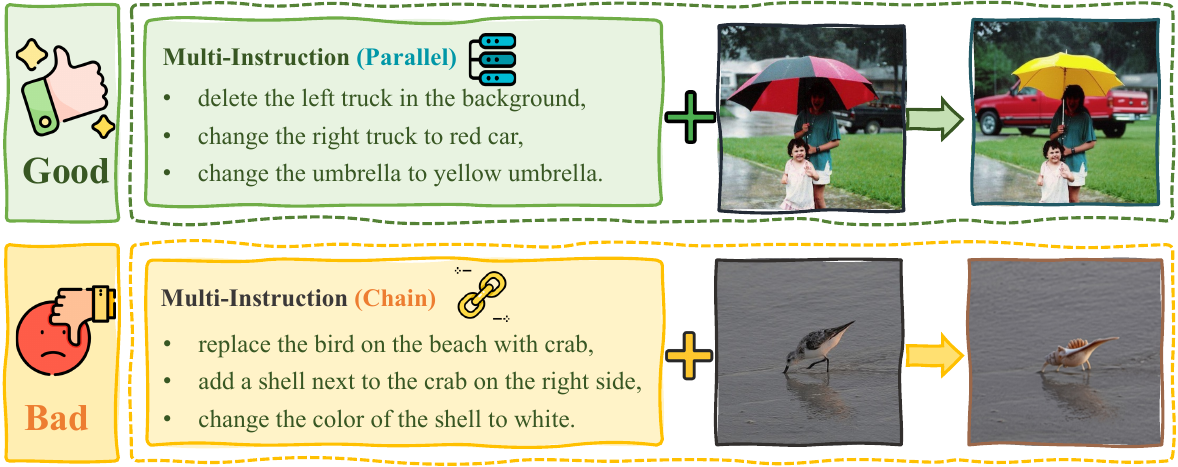}
    \vspace{-7mm}
    \caption{\small Comparison between parallel and chain multi-instruction image editing. \textbf{Parallel editing} applies independent instructions simultaneously, while \textbf{chain editing} involves dependent instructions that must be executed in sequence.}
    \label{fig:intro}
    \vspace{-3mm}
\end{figure}

However, real-world editing often demands more than singular directives. Users frequently issue complex instructions comprising multiple, sometimes interdependent, sub-tasks forming a ``chain'' where one step's outcome affects the next (Figure~\ref{fig:intro}). While current models adeptly handle isolated edits, their ability to understand, decompose, and execute such multi-faceted, particularly chain-instruction, inputs is a significant challenge, often leading to incoherent results or failure to meet all constraints.

This limitation stems partly from prevalent evaluation benchmarks. As Table~\ref{tab:compare} shows, most existing benchmarks primarily test single-constraint or simple independent multi-instruction edits, lacking systematic evaluation for complex, dependent instructions. For example, MagicBrush~\cite{zhang2023magicbrush} and $\text{I}^2\text{EBench}$~\cite{I2ebench2024}, while instruction-driven, overlook multi-instruction and chain-instruction complexities. Complex-Edit~\cite{yang2025texttt} addresses multi-instruction editing but not specifically chained dependencies, and its LLM-based consistency evaluation struggles to assess non-modified regions. Furthermore, common L1/L2 consistency metrics are flawed, incorrectly favoring unedited images.

To propel image editing models towards a higher level of intelligence, they must learn ``combinatorial reasoning'', the ability to execute instructions correctly under the premise that multiple constraints must hold true simultaneously, especially when these constraints are sequentially dependent. To address the aforementioned limitations and better evaluate the instruction-following capabilities of existing editing models in complex scenarios, this paper introduces \textbf{ComplexBench-Edit}, a novel benchmark for image editing specifically designed to assess performance on complex instructions involving multiple combined and dependent modifications. Our benchmark systematically evaluates how well models can handle both parallel and, critically, chain-dependent instructions. Furthermore, we propose a novel vision consistency evaluation method that excludes the influence of modified content by assessing consistency only in the remaining, unaltered regions. We also introduce a simple yet powerful CoT-based approach for image editing.
The main contributions of this work are:
\begin{itemize}[leftmargin=*]
    \item We propose ComplexBench-Edit, a new benchmark tailored for evaluating image editing models on complex, multi-instruction, and chain-dependent instructions, along with a novel vision consistency metric.
    \item We provide a detailed analysis of current state-of-the-art image editing models on ComplexBench-Edit, highlighting their strengths and weaknesses in handling complex directives.
    \item We introduce a simple yet effective CoT-based approach to enhance the complex instruction following capabilities of existing models, demonstrating its effectiveness on our benchmark.
\end{itemize}

\begin{table}[t]\footnotesize
\centering
\caption{\small ``Ins.'', ``Multi-ins.'', ``Chain-ins.'', and ``Pixel-eval'' denote ``Instruction-driven'', ``Multi-instruction'', ``Chain-instruction'', and ``Pixel-level image eval'', respectively.}
\label{tab:compare}
\vspace{-3mm}
\begin{tabular}{@{}lccccc@{}}
\toprule
\textbf{Datasets / Benchmarks} & \textbf{Ins.} & \textbf{Multi-ins.} & \textbf{Chain-ins.} & \textbf{Pixel-eval} &  \\ \midrule
MagicBrush~\cite{zhang2023magicbrush} & \color{teal}\checkmark & $\color{red}\times$ & $\color{red}\times$ & \color{teal}\checkmark &  \\
$\textbf{I}^2$EBench~\cite{I2ebench2024} & \color{teal}\checkmark & $\color{red}\times$  & $\color{red}\times$ & \color{teal}\checkmark &  \\
AnyEdit~\cite{yu2024anyedit} & \color{teal}\checkmark & $\color{red}\times$ & $\color{red}\times$  & \color{teal}\checkmark &  \\
UltraEdit~\cite{zhao2024ultraedit} & \color{teal}\checkmark & $\color{red}\times$ & $\color{red}\times$  & \color{teal}\checkmark  &  \\
PIE-Bench++~\cite{huang2024paralleledits} & $\color{red}\times$  & $\color{red}\times$  & $\color{red}\times$  & \color{teal}\checkmark &  \\
Complex-Edit~\cite{yang2025texttt} & \color{teal}\checkmark & \color{teal}\checkmark & $\color{red}\times$  &  $\color{red}\times$  &  \\
Ours (ComplexBench-Edit) & \color{teal}\checkmark & \color{teal}\checkmark & \color{teal}\checkmark & \color{teal}\checkmark &  \\ \bottomrule
\end{tabular}
\vspace{-3mm}
\end{table}

\section{Related Work}
\subsection{Text-driven Image Editing}
Diffusion Models (DMs)~\cite{ddpm,d3pm,wang2024diffusion, song2020score,podell2023sdxl} have significantly advanced image generation and editing, leading to the Instruction-Based Image Editing (IIE) task where models alter images based on textual instructions. Early IIE milestones include InstructPix2Pix~\cite{brooks2023instructpix2pix}, which was trained on synthetic data from GPT-3~\cite{brown2020language} and Prompt-to-Prompt~\cite{hertz2022prompt}. Performance was further improved by works like MagicBrush~\cite{zhang2023magicbrush}, UltraEdit~\cite{zhao2024ultraedit}, SEED-Data-Edit~\cite{ge2024seed}, HumanEdit~\cite{bai2024humanedit}, and AnyEdit~\cite{yu2024anyedit}, which utilized high-quality curated datasets.
More recently, Large Language Models (LLMs)~\cite{zhou2025weak,zhou2024visual} have been integrated to enhance instruction comprehension. For instance, MGIE~\cite{MGIE2024} and SmartEdit~\cite{huang2024smartedit} leverage Multimodal Large Language Models (MLLMs) for precise guidance. OmniGen~\cite{xiao2024omnigen} employs Phi-3~\cite{abdin2024phi} to strengthen instruction understanding, while Step1X-Edit~\cite{liu2025step1x} uses MLLMs~\cite{bai2025qwen2} for accurate parsing and fine-tunes on quality datasets. ICEdit~\cite{zhang2025context} adopts DiT-based models~\cite{peebles2023scalable} for their strong generative power. Despite excellent performance on single-instruction tasks, the ability of these models to process and understand complex multi-instruction or chained instructions remains largely unevaluated.

\subsection{Image Editing Benchmarks}
Various benchmarks have recently emerged to advance image editing. MagicBrush~\cite{zhang2023magicbrush}, based on COCO~\cite{lin2014microsoft}, evaluates instruction following and consistency using metrics like CLIP~\cite{radford2021learning}, DINO~\cite{zhang2022dino}, L1, and L2 distances. $\mathbf{I}^2$EBench~\cite{I2ebench2024} expanded editing types with 16 evaluation dimensions for comprehensive assessment. Prompt-based benchmarks like PIE-Bench++~\cite{huang2024paralleledits}, OIR-Bench~\cite{yang2023object}, and LOMOE-Bench~\cite{chakrabarty2024lomoe} evaluate multi-object editing but differ from instruction-based methods by requiring image-caption pairs.
While these benchmarks have driven progress, they mainly focus on single-instruction editing or specific descriptive inputs. Complex-Edit~\cite{yang2025texttt} addresses complex instructions by integrating atomic tasks via a ``Chain-of-Edit'' pipeline, but it overlooks the specific challenges of chain-instruction scenarios. Therefore, a systematic benchmark for detailed evaluation in complex, particularly chained, scenarios is crucial.

\section{ComplexBench-Edit}
\begin{figure*}
    \centering
    \includegraphics[width=1\linewidth]{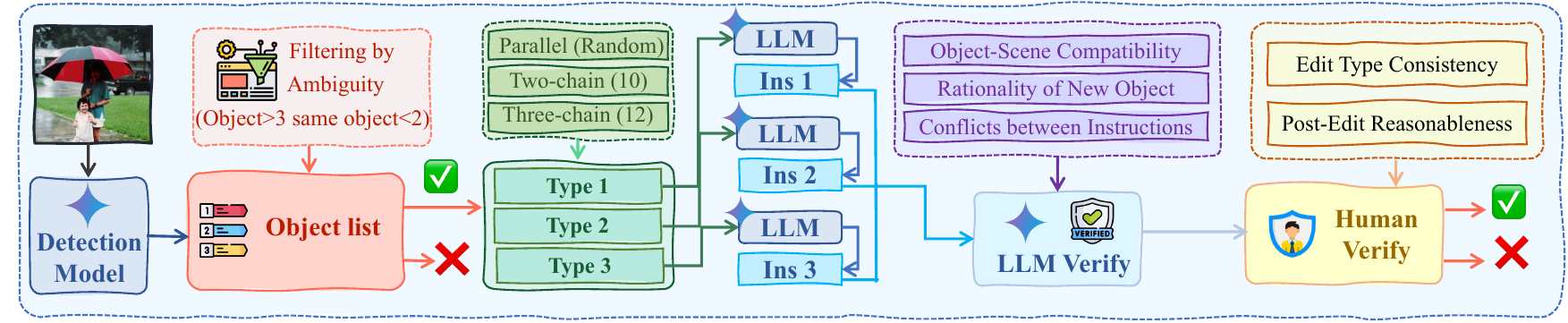}
    \vspace{-7mm}
    \caption{\small Overview of the data creation pipeline of ComplexBench-Edit.}
    \label{fig:pipeline}
    \vspace{-3mm}
\end{figure*}

We will detail the construction of our ComplexBench-Edit dataset. The overall data creation pipeline is shown in Figure~\ref{fig:pipeline}.
\subsection{Dataset Construction}

\subsubsection{Vision Content Filter}
The Vision Content Filter represents the initial and essential stage in our dataset construction pipeline. Its primary objective is to select source images that are suitable for generating complex editing instructions, based on an analysis of their visual content. A fundamental requirement for defining object-centric editing tasks is understanding the objects present in an image. This requires applying an object detection model to identify and localize objects. However, since our source dataset, MSCOCO \cite{lin2014microsoft}, provides rich ground-truth annotations, we leverage these directly, bypassing the need for a separate detection process. Specifically, we use the annotated bounding boxes, which accurately indicate both the class and location of each object within an image $I$. We introduce two filtering criteria:

\textbf{1. Intra-class Frequency Constraint.}
To reduce ambiguity when referring to multiple instances of the same object category in textual instructions, we restrict the number of objects per class. Let $N\_c(I)$ denote the number of instances (i.e., bounding boxes) of class $c$ in image $I$. We discard image $I$ if there exists any class $c$ such that:
\begin{align}
\exists c \in C(I), \quad N\_c(I) > 2
\end{align}
This ensures that no object category appears more than twice in any selected image, simplifying unambiguous reference.

\textbf{2. Category Diversity Constraint.}
To ensure visual complexity and semantic richness—key for generating multi-object editing instructions—we require a minimum level of object diversity. Let $C(I)$ be the set of unique object categories present in image $I$. We discard image $I$ if:
\begin{align}
|C(I)| < 3
\end{align}
This guarantees that each selected image contains at least three distinct object categories.

An image $I$ passes the Vision Content Filter if and only if the following two conditions are satisfied:
\begin{align}
\forall c \in C(I),\quad N_c(I) \leq 2  ~\land~ |C(I)| \geq 3
\end{align}

\subsubsection{Type-guided Instruction Generation}
\begin{figure}
    \centering
    \includegraphics[width=0.65\linewidth]{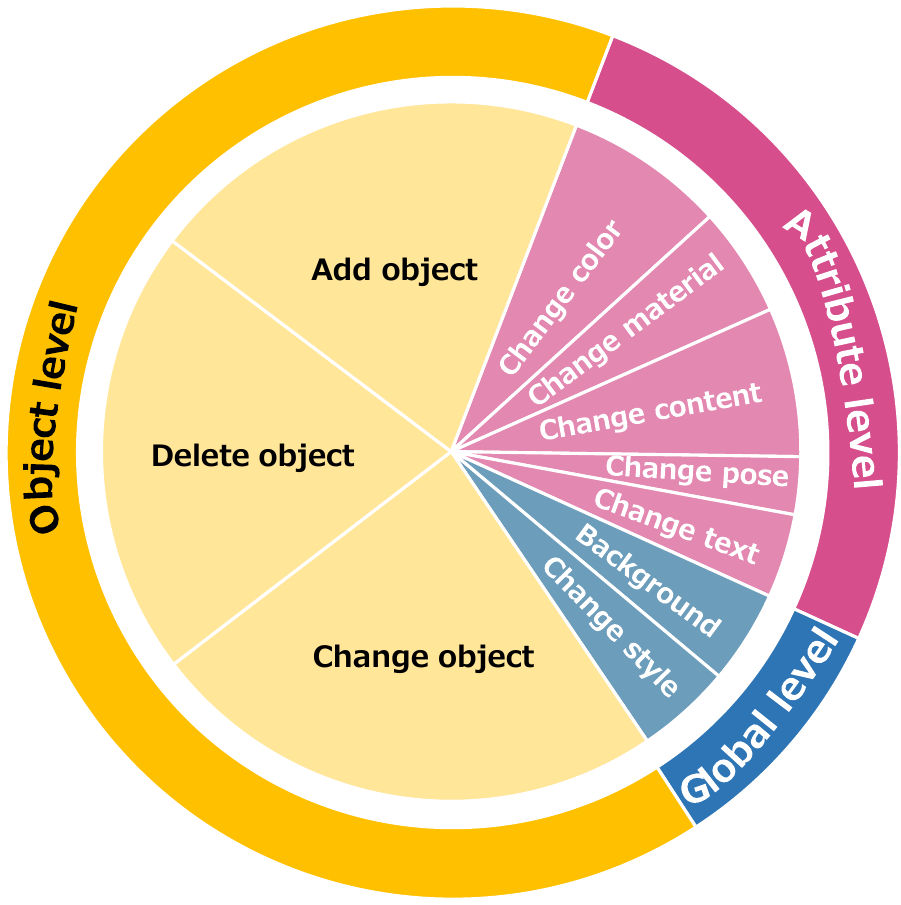}
    \vspace{-3mm}
    \caption{\small Overview of the three hierarchical levels of editing types in ComplexBench-Edit.}
    \label{fig:type_levels}
    \vspace{-3mm}
\end{figure}

After filtering source images based on visual content, the subsequent step involves generating complex editing instructions. This stage translates curated image data and predefined editing objectives into concrete, executable textual commands for image manipulation. The core mechanism is a MLLM, which takes as input the selected image $I$, its object list $\text{obj}(I)$, and a chosen combination of editing types. Here, $\text{obj}(I)$ denotes the list of objects detected in image $I$, extracted by a prior object detection module. The MLLM analyzes the image and object-level information in conjunction with the specified editing types, determines appropriate target objects for each operation, and produces precise textual instructions. The structure and complexity of these instructions are governed by predefined editing type combinations, categorized into three hierarchical levels, as shown in Figure~\ref{fig:type_levels}.

\ding{182} \textit{Level 1: Parallel.} This level consists of three editing instructions that are logically independent. The order of execution does not affect the final outcome. All three instruction types are randomly sampled from a general set of 10 predefined editing operations.

\ding{183} \textit{Level 2: Two-chain.} This level includes a two-step dependency chain and one additional independent instruction. Let the instruction types be denoted as $T_1, T_2, T_3$, where $T_1 \rightarrow T_2$ forms a logical dependency and $T_3$ is independent:
\begin{align}
\text{Dependent chain:} \quad & T_1 \rightarrow T_2 \\
\text{Independent instruction:} \quad & T_3
\end{align}
Types $T_1$ and $T_2$ are sampled from a curated subset of 10 editing types to ensure logical consistency and feasibility (e.g., avoiding cases where an object is modified after being deleted). The independent type $T_3$ is sampled from the general editing type set.

\ding{184} \textit{Level 3: Three-chain.} This level involves a sequence of three interdependent instructions, forming a three-step chain:
\begin{align}
T_1 \rightarrow T_2 \rightarrow T_3
\end{align}
All types in this level are sampled from a dedicated subset of 12 editing operations designed to support deeper, logically coherent dependencies.

For levels involving dependent instructions (Two-chain and Three-chain), generation is performed sequentially. When producing an instruction that depends on a previous one, the MLLM receives the previously generated instruction text as additional context:
\begin{align}
\text{Input}_i = \{ I,\; \text{obj}(I),\; T_i,\; T_{i-1} \}
\end{align}
This context-aware generation process ensures logical consistency and coherence across dependent editing operations.

\subsubsection{Editing Instruction Feasibility Check.}
% Gemini for editing instruction evaluation
Following the generation of editing instructions by the MLLM, a crucial step is to validate their feasibility. The complexity of multi-step operations and potential semantic inconsistencies require a rigorous check. We employ a separate validating MLLM that analyzes the generated instructions alongside the source image and object information. A sample (image + instructions) is deemed valid only if it passes checks based on the following criteria:

\textbf{1. Object-Scene Compatibility.}
Ensures that objects targeted for placement, addition, or movement are semantically consistent and visually plausible within the image's scene context (e.g., adding a fish to water is compatible; adding it to the sky is not). This prevents instructions that result in unrealistic or nonsensical scenarios.

\textbf{2. Rationality of Object Reference and Attributes.}
Verifies that references to existing objects are unambiguous. For added or modified objects, it checks if the specified object type and attributes are rational and non-contradictory (e.g., a 'flying car' might be rational in some contexts, but 'invisible, heavy box' might not be).

\textbf{3. Conflicts between Instructions.}
For samples with multiple instructions, this criterion identifies potential logical conflicts. It checks if instructions contradict each other or if a step renders a subsequent dependent instruction infeasible (e.g., deleting an object required for a later modification).

Samples that fail any of these feasibility checks are discarded. This validation ensures that the ComplexBench-Edit dataset contains high-quality, executable instructions suitable for benchmarking complex image editing capabilities.

\subsubsection{Human Review.}
% human for editing instructions available
Despite the MLLM-based feasibility check, a crucial human review stage is performed for benchmark quality assurance. Two PhD students reviewed a subset of generated samples (image + instructions) based on two dimensions:

\textbf{1. Edit Type Consistency.}
Ensures the generated instruction text accurately reflects the intended editing operation type and category (e.g., Parallel, Two-chain, specific action). This aligns the instruction with the predefined task structure.

\textbf{2. Post-Edit Reasonableness.}
Evaluates the semantic and visual plausibility of the scene after hypothetically executing the instruction(s). Checks if implied changes or additions are reasonable and coherent within the original image's context, assessing the logical soundness of the envisioned modified scene.

\subsection{Aumated Evaluation Metrics}
To quantitatively evaluate image editing models on ComplexBench-Edit, we define a set of automated evaluation metrics. These metrics assess both the model's ability to correctly execute complex, multi-step editing instructions and the visual quality and consistency of the generated results. 
\subsubsection{Editing Performance Evaluation.}
Editing performance is quantitatively evaluated based on the model's ability to accurately execute both single and chain-dependent instructions. This assessment is performed automatically by a dedicated MLLM evaluator. For each instruction instance, a score is assigned on a 5-point scale (0-5), reflecting the quality of the generated image corresponding to that specific edit. A score of 5 indicates perfect execution, while 0 indicates complete failure or no perceivable attempt. The MLLM evaluator assigns this score by evaluating the edited image against a predefined set of type-specific criteria. 
% These criteria assess aspects critical to the success of each editing operation, such as spatial accuracy, semantic correctness, realism, integration with the original image, and absence of artifacts.

\textbf{1. Single-Instruction Evaluation}
For independent instructions, the MLLM evaluator assigns a single 5-point score based on the defined criteria for its specific editing type.

\textbf{2. Chain-Instruction Evaluation}
For instructions forming a dependent chain, the evaluation methodology accounts for the prerequisite nature of earlier steps for the successful execution of subsequent ones. The MLLM evaluator assigns a 5-point score ($S_i$) to each individual instruction $i$ within the chain, based on its type-specific criteria. The overall performance score for a chain of $N$ instructions is calculated as the product of the scores of its constituent instructions:
\begin{align}
S_{chain} = S_1 \times S_2 \times \dots \times S_N
\end{align}
This multiplicative approach ensures that failure in any single step (resulting in $S_i=0$) propagates to the entire chain, yielding a total score of 0 and accurately reflecting the inability to complete the dependent task sequence successfully.

\subsubsection{Vision Consistency Evaluation.}
Beyond assessing the performance of the targeted editing operations, it is equally important to evaluate the model's ability to preserve the original content and visual consistency in the regions of the image that were not intended for modification. Standard pixel-wise metrics like L1 or L2 distance computed over the entire image are inadequate for this purpose, as they would erroneously assign the best scores to models that return the original, unedited image (resulting in a distance of zero), failing to reflect any editing capability.

To address this, we propose a region-specific consistency evaluation. We identify the regions potentially affected by the editing process and exclude them from the consistency calculation. Specifically, we leverage the bounding boxes associated with objects. We consider the ground truth bounding boxes from the original image (available from MSCOCO annotations) and the bounding boxes detected in the edited image using a robust object detection model. Let $B_{orig}$ be the set of ground truth bounding boxes in the image $I_{orig}$, and $B_{edit}$ be the set of bounding boxes detected in the edited image $I_{edit}$. We define the set of regions potentially subject to modification as the union of the areas covered by bounding boxes:
\begin{align}
R_{edit} = \bigcup_{b \in B_{orig} \cup B_{edit}} \text{Area}(b)
\end{align}
where $\text{Area}(b)$ denotes the region covered by bounding box $b$.

The vision consistency is then evaluated on the complementary region, $R_{consistent} = \text{Image Area} \setminus R_{edit}$, which represents the parts of the image that should ideally remain unchanged. We compute the consistency using standard pixel-wise distance metrics, specifically the L1 distance and L2 distance, applied only to the pixels within the region $R_{consistent}$:
\begin{align}
\text{L1} &= \frac{1}{|R_{consistent}|} \sum_{p \in R_{consistent}} |I_{orig}(p) - I_{edit}(p)| \\
\text{L2} &= \frac{1}{|R_{consistent}|} \sum_{p \in R_{consistent}} (I_{orig}(p) - I_{edit}(p))^2
\end{align}
where $|R_{consistent}|$ is the number of pixels in the region $R_{consistent}$, and $I(p)$ denotes the pixel value at location $p$. This approach ensures that our consistency metric specifically measures the preservation of the background and non-target content, penalizing unintended modifications or artifacts outside the designated editing areas, without unfairly rewarding models that perform no edits.

\subsection{CoT Reasoning for Image Editing}
\begin{figure}[t]
    \centering
    \includegraphics[width=1\linewidth]{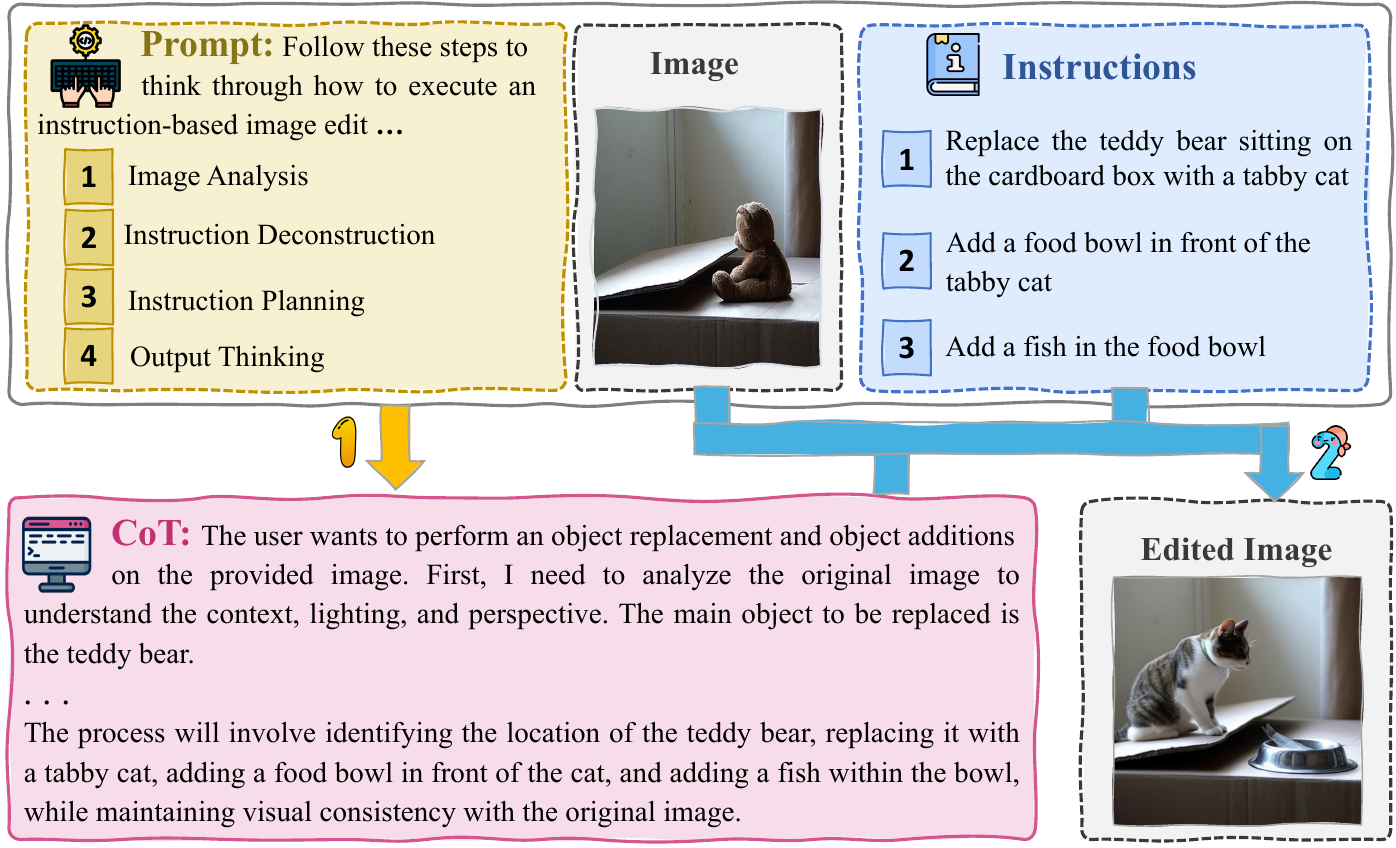}
    \vspace{-7mm}
    \caption{\small Diagram of the proposed Chain-of-Thought (CoT) reasoning approach for image editing.}
    \label{fig:agent}
    \vspace{-3mm}
\end{figure}
In addition to introducing a new benchmark and automated evaluation metrics, we propose a powerful training-free baseline leveraging Chain-of-Thought (CoT) reasoning. Inspired by recent work~\cite{guo2025can, mitra2024compositional, wei2022chain} showing that MLLM-generated CoT enhances instruction understanding in image generation, we apply this to image editing.

As shown in Figure~\ref{fig:agent}, our baseline employs a separate MLLM to generate a detailed CoT rationale for executing a given instruction on the source image. A specific prompt is used to guide the MLLM's thought process, encouraging it to: 1) Analyze the image context, 2) Deconstruct the user instruction into specific operations, 3) Plan the spatial and sequential execution of these operations, and 4) Construct a conceptual output or execution blueprint. 

The generated CoT is then concatenated with the original user instruction. This combined input string is provided to the image editing model. This approach aims to provide the editing model with a richer, reasoned understanding of the task, facilitating improved performance without requiring task-specific fine-tuning.

\section{Experiments}
\subsection{Comparison Methods}
We evaluate the performance on ComplexBench-Edit against a comprehensive suite of recent image editing models.
These include established instruction-based methods like InstructPix2Pix~\cite{brooks2023instructpix2pix}, MagicBrush~\cite{zhang2023magicbrush}, UltraEdit~\cite{zhao2024ultraedit}, ICEdit~\cite{zhang2025context}, and AnyEdit~\cite{yu2024anyedit}.
We also benchmark against models that integrate LLMs for improved instruction comprehension and editing, such as SEED-LLAMA~\cite{seed-llama}, OmniGen~\cite{xiao2024omnigen}, and Step1X-Edit~\cite{liu2025step1x}.
Additionally, we include VAR-GPT~\cite{zhuang2025vargpt} and GoT~\cite{fang2025got}.
Finally, we use the powerful proprietary MLLM, Gemini~\cite{team2023gemini}, as a strong baseline editor, and compare it with our proposed Gemini-CoT method, which augments Gemini with Chain-of-Thought reasoning.

\subsection{Editing Performance Evaluation}
\textbf{Evaluation with Different Complex-levels}
Table~\ref{table1} shows performance across varying instruction complexities. Our \textbf{Gemini-CoT} method consistently achieves the highest scores, excelling at Parallel, Two-chain, and Three-chain instructions, with CoT reasoning providing a clear advantage over the strong standard Gemini. Performance generally degrades with increased complexity, as most models struggle significantly with chained dependencies. While recent models like Step1X-Edit and ICEdit show improvements, they lag behind Gemini-based approaches, especially in longer dependency chains where Gemini-CoT leads substantially. It shows challenge of sequential edits and efficacy of our method.

\begin{table}[!t]\footnotesize
\centering
\caption{\small Performance comparison on ComplexBench-Edit across different instruction complexity levels: Parallel (Obj., Obj.-At., Obj.-At.-G.), Two-chain (Obj.-At.), and Three-chain (Obj.-At.). ``Obj.'' denotes Object, ``At.'' denotes Attribute, and ``G.'' denotes Global.}
\label{table1}
\vspace{-3mm}
\setlength{\tabcolsep}{1.5pt}
\resizebox{\linewidth}{!}{
\begin{tabular}{lcccccc}
\toprule
\multicolumn{1}{c}{} &
  \multicolumn{3}{c}{\textbf{Parallel}} &
  \multicolumn{1}{c}{\textbf{Two-chain}} &
  \multicolumn{1}{c}{\textbf{Three-chain}} &
  \multicolumn{1}{c}{}\\ \cmidrule(lr){2-4} \cmidrule(lr){5-5} \cmidrule(lr){6-6}  
\multicolumn{1}{c}{\multirow{-2}{*}{\textbf{Model}}} &
  \multicolumn{1}{c}{\textbf{Obj.}} &        % Abbreviated
  \multicolumn{1}{c}{\textbf{Obj.-At.}} &  % Abbreviated
  \multicolumn{1}{c}{\textbf{Obj.-At.-G.}} & % Abbreviated
  \multicolumn{1}{c}{\textbf{Obj.-At.}} &  % Abbreviated (already fits common understanding)
  \multicolumn{1}{c}{\textbf{Obj.-At.}} &  % Abbreviated (already fits common understanding)
  \multicolumn{1}{c}{\multirow{-2}{*}{\textbf{Avg.}}}\\ \midrule
\textbf{InstructPix2Pix}~\cite{brooks2023instructpix2pix} & ~~~~~~9.95        & 10.45       & 13.47       & ~~~~~~4.83        & ~~~~~~0.52        & ~~~~~~7.85       \\
\textbf{MagicBrush}~\cite{zhang2023magicbrush}      & 18.29       & 14.65       & 11.90       & ~~~~~~9.70        & ~~~~~~0.57 & 11.02       \\
\textbf{UltraEdit}~\cite{zhao2024ultraedit}       & 24.46       & 25.78       & 23.80       & 13.74       & ~~~~~~3.10 & 18.18       \\
\textbf{AnyEdit}~\cite{yu2024anyedit}         & 13.51       & 11.27       & ~~~~~~9.33        & ~~~~~~6.61        & ~~~~~~1.01 & ~~~~~8.35       \\
\textbf{VAR-GPT}~\cite{zhuang2025vargpt}         & ~~~~~~0.86        & ~~~~~~0.39        & ~~~~~~3.80        & ~~~~~~0.54        & ~~~~~~0.00 &~~~~~1.12       \\
\textbf{SEED-LLAMA}~\cite{seed-llama}      & ~~~~~~5.59        & ~~~~~~5.93        & ~~~~~~6.00        & ~~~~~4.48        & ~~~~~~0.27 &~~~~~4.46      \\
\textbf{OmniGen}~\cite{xiao2024omnigen}         & 32.34       & 26.57       & 30.90       & 16.96       & ~~~~~~4.36 & 22.23       \\
\textbf{GoT}~\cite{fang2025got} & 20.09 & 14.85 & 16.42 & 11.76 & ~~~~~0.93 & 12.81  \\
\textbf{ICEdit}~\cite{zhang2025context}          &  32.97 & 26.96       & 27.07       &  21.80     & ~~~~~8.09 & 23.38 \\
\textbf{Step1X-Edit}~\cite{liu2025step1x}     & 32.61       &  30.14 &  31.57 & 19.56 & ~~~~~~6.34 & 24.05       \\
\textbf{Gemini}~\cite{team2023gemini} &
  {\ul 49.31} &
  {\ul 43.47} &
  {\ul 37.23} &
  {\ul 38.35} &
  {\ul 15.10} & {\ul 36.70} \\
\rowcolor[HTML]{ECF4FF}
\textbf{Ours~(Gemini-CoT)} &
\textbf{51.76} &
\textbf{50.11} &
\textbf{43.08} &
\textbf{39.85} &
\textbf{17.54} & \textbf{40.47}\\
\bottomrule  
\end{tabular}
}
\vspace{-3mm}
\end{table}

\noindent\textbf{Evaluation with Different Complex-types}
Table~\ref{table2} details performance across 10 distinct editing types. \textbf{Gemini-CoT} again shows broad superiority, leading across most object, attribute, and global-level edits. Vanilla Gemini also excels, with CoT notably boosting performance, particularly for complex object and attribute modifications. While some models show niche strengths (e.g., Step1X-Edit in ``Change Style''), many struggle with types like ``Change Pose'' or ``Change Text'' compared to Gemini variants. Results highlight MLLM approaches like Gemini and our Gemini-CoT set benchmark.

\begin{table*}[!t]\small
\centering
\caption{\small Performance comparison of models across 10 distinct editing types, categorized into Object-level, Attribute-level, and Global-level. }
\label{table2}
\vspace{-1mm}
\setlength{\tabcolsep}{6.6pt}
\begin{tabular}{lcccccccccc}
\toprule 
\multicolumn{1}{c}{} & \multicolumn{3}{c}{\textbf{Object-level}} & \multicolumn{5}{c}{\textbf{Attribute-level}} & \multicolumn{2}{c}{\textbf{Global-level}} \\ \cmidrule(lr){2-4} \cmidrule(lr){5-9} \cmidrule(lr){10-11}  
\multicolumn{1}{c}{\multirow{-2}{*}{\textbf{Model}}} & \textbf{\begin{tabular}[c]{@{}c@{}}Add\\ Object\end{tabular}} & \textbf{\begin{tabular}[c]{@{}c@{}}Change\\ Object\end{tabular}} & \textbf{\begin{tabular}[c]{@{}c@{}}Delete\\ Object\end{tabular}} & \textbf{\begin{tabular}[c]{@{}c@{}}Change\\ Color\end{tabular}} & \textbf{\begin{tabular}[c]{@{}c@{}}Change\\ Pose\end{tabular}} & \textbf{\begin{tabular}[c]{@{}c@{}}Change\\ Material\end{tabular}} & \textbf{\begin{tabular}[c]{@{}c@{}}Change\\ Content\end{tabular}} & \textbf{\begin{tabular}[c]{@{}c@{}}Change\\ Text\end{tabular}} & \textbf{\begin{tabular}[c]{@{}c@{}}Change\\ Background\end{tabular}} & \textbf{\begin{tabular}[c]{@{}c@{}}Change\\ Style\end{tabular}} \\ \midrule
\textbf{InstructPix2Pix}~\cite{brooks2023instructpix2pix} & 11.10 & 10.51 & 12.50 & 13.98 & ~~~~~9.09 & ~~~~~9.38 & 12.82 & ~~~~~2.11 & ~~~~~9.70 & 18.02 \\
\textbf{MagicBrush}~\cite{zhang2023magicbrush} & 13.96 & 14.06 & 22.97 & 11.84 & ~~~~~9.55 & ~~~~~9.07 & 13.94 & ~~~~~7.11 & 10.91 & ~~~~~4.36 \\
\textbf{UltraEdit}~\cite{zhao2024ultraedit} & 24.63 & 24.12 & 29.59 & 24.20 & ~~~~~9.52 & 17.11 & 29.22 & 17.63 &  19.59 & 25.35 \\
\textbf{AnyEdit}~\cite{yu2024anyedit} & ~~~~~9.90 & ~~~~~8.83 & 16.83 & 20.00 & ~~~~~5.91 & ~~~~~5.57 & 13.66 & ~~~~~8.42 & ~~~~~6.67 & ~~~~~0.99 \\
\textbf{VAR-GPT}~\cite{zhuang2025vargpt} & ~~~~~1.77 & ~~~~~0.51 & ~~~~~4.27 & ~~~~~0.00 & ~~~~~1.82 & ~~~~~0.41 & ~~~~~0.56 & ~~~~~0.00 & ~~~~~0.40 & ~~~~~3.37 \\
\textbf{SEED-LLAMA}~\cite{seed-llama} & ~~~~~5.31 & ~~~~~5.48 & ~~~~~8.82 & ~~~~~2.91 & ~~~~~8.16 & ~~~~~5.57 & ~~~~~3.10 & ~~~~~0.79 & ~~~~~5.66 & ~~~~~6.34 \\
\textbf{OmniGen}~\cite{xiao2024omnigen} & 22.83 & 26.29 &  42.45 & 27.77 & ~~~~~8.18 & 15.00 & 33.66 & 16.84 & 17.17 & {\ul 44.75} \\
\textbf{GoT}~\cite{fang2025got} & 15.18 & 15.18 & 26.73 & ~~~~~9.32 & ~~~~~6.82 & 11.13 & 11.69 & ~~~~~2.89 & 11.52 & 25.80 \\
\textbf{ICEdit}~\cite{zhang2025context} & 30.31 &  31.22 & 26.52 & {\ul 41.55} & ~~~~~7.73 & 16.08 & 34.93 & 17.37 & 17.38 & 42.18 \\
\textbf{Step1X-Edit}~\cite{liu2025step1x} &  32.29 & 30.91 & 33.48 & 34.56 &  10.00 & {\ul 22.06} &  38.45 &  25.26 & 16.97 & \cellcolor[HTML]{ECF4FF}\textbf{45.54} \\
\textbf{Gemini}~\cite{team2023gemini} & {\ul 47.87} & {\ul 42.63} & {\ul 56.99} & 36.38 & \cellcolor[HTML]{ECF4FF}\textbf{52.38} &  19.78 & \cellcolor[HTML]{ECF4FF}\textbf{46.07} & {\ul 26.93} & {\ul 31.22} & 16.97 \\
\textbf{Ours (Gemini-CoT)} & \cellcolor[HTML]{ECF4FF}\textbf{48.23} & \cellcolor[HTML]{ECF4FF}\textbf{44.78} & \cellcolor[HTML]{ECF4FF}\textbf{66.38} & \cellcolor[HTML]{ECF4FF}\textbf{41.98} & {\ul 51.36} & \cellcolor[HTML]{ECF4FF}\textbf{32.50} & {\ul 45.82} & \cellcolor[HTML]{ECF4FF}\textbf{28.68} & \cellcolor[HTML]{ECF4FF}\textbf{38.98} & 30.69\\
\bottomrule
\end{tabular}
\vspace{-1mm}
\end{table*}

\begin{figure}[t]
    \centering
    \includegraphics[width=0.95\linewidth]{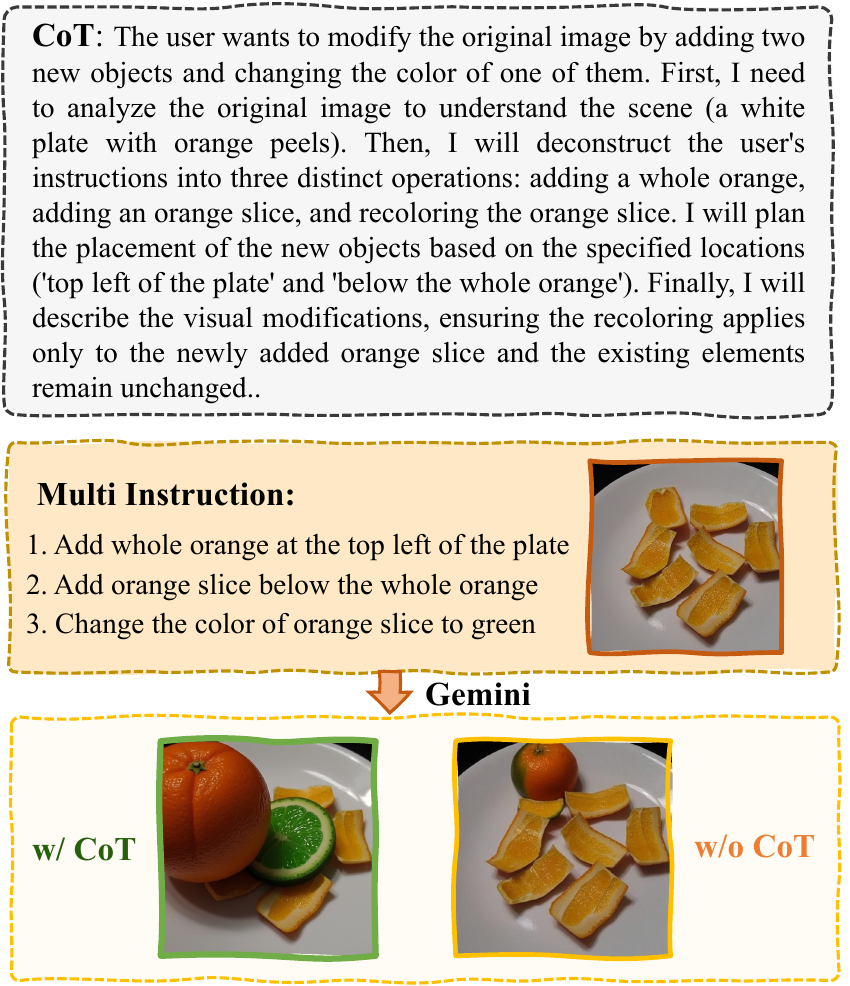}
    \vspace{-3mm}
    \caption{\small Comparison of image editing results w/ and w/o CoT.}
    \label{fig:case}
    \vspace{-3mm}
\end{figure}

\begin{table*}[!t]\small
\centering
\caption{\small Vision consistency evaluation (L1 and L2 distances, lower is better) on non-edited regions across instruction complexities. }
\label{table3}
\vspace{-1mm}
\setlength{\tabcolsep}{8pt}
\begin{tabular}{lcccccccccc}
\toprule
\multicolumn{1}{c}{} & \multicolumn{4}{c}{\textbf{Parallel}} & \multicolumn{2}{c}{\textbf{Two-chain}} & \multicolumn{2}{c}{\textbf{Three-chain}} &\multicolumn{2}{c}{}  \\ \cmidrule(lr){2-5} \cmidrule(lr){6-7} \cmidrule(lr){8-9} 
\multicolumn{1}{c}{} & \multicolumn{2}{c}{\textbf{Object}} & \multicolumn{2}{c}{\textbf{Object-attribute}} & \multicolumn{2}{c}{\textbf{Object-attribute}} & \multicolumn{2}{c}{\textbf{Object-attribute}} & \multicolumn{2}{c}{\multirow{-2}{*}{\textbf{Avg.}}}
\\ \cmidrule(lr){2-3} \cmidrule(lr){4-5} \cmidrule(lr){6-7} \cmidrule(lr){8-9} \cmidrule(lr){10-11}  
\multicolumn{1}{c}{\multirow{-3}{*}{\textbf{Model}}} & \multicolumn{1}{c}{\textbf{L1 ↓}} & \multicolumn{1}{c}{\textbf{L2 ↓}} & \multicolumn{1}{c}{\textbf{L1 ↓}} & \multicolumn{1}{c}{\textbf{L2 ↓}} & \multicolumn{1}{c}{\textbf{L1 ↓}} & \multicolumn{1}{c}{\textbf{L2 ↓}} & \multicolumn{1}{c}{\textbf{L1 ↓}} & \multicolumn{1}{c}{\textbf{L2 ↓}}  &\multicolumn{1}{c}{\textbf{L1 ↓}} & \multicolumn{1}{c}{\textbf{L2 ↓}} \\ \cmidrule{1-11}
\textbf{InstructPix2Pix}~\cite{brooks2023instructpix2pix} & 0.1663 & 0.0588 & 0.1581 & 0.0560 & 0.1696 & 0.0617 & 0.1374 & 0.0428 & 0.1578 & 0.0548 \\
\textbf{MagicBrush}~\cite{zhang2023magicbrush} & 0.0701 & 0.0233 & 0.0707 & 0.0242 & 0.0609 & 0.0210 & 0.0599 & 0.0180 & 0.0654 & 0.0216 \\
\textbf{UltraEdit}~\cite{zhao2024ultraedit} & 0.0568 & 0.0098 & 0.0522 & {\ul 0.0081} & 0.0556 & {\ul 0.0092} & 0.0596 & 0.0099& 0.0560 & {\ul 0.0092} \\
\textbf{AnyEdit}~\cite{yu2024anyedit} & 0.0685 & 0.0239 & 0.0602 & 0.0194 & 0.0747 & 0.0292 & 0.0424 & {\ul 0.0078} & 0.0614 & 0.0201 \\
\textbf{VAR-GPT}~\cite{zhuang2025vargpt} & 0.2539 & 0.1165 & 0.2421 & 0.1112 & 0.2400 & 0.1066 & 0.2390 & 0.1038 & 0.2437 & 0.1095 \\
\textbf{SEED-LLAMA}~\cite{seed-llama} & 0.2763 & 0.1223 & 0.2754 & 0.1213 & 0.2716 & 0.1184 & 0.2727 & 0.1174 & 0.2740 & 0.1198 \\
\rowcolor[HTML]{ECF4FF} 
\textbf{OmniGen}~\cite{xiao2024omnigen} & \textbf{0.0392} & \textbf{0.0071} & \textbf{0.0352} & \textbf{0.0059} & \textbf{0.0381} & \textbf{0.0079} & \textbf{0.0382} & \textbf{0.0062} & \textbf{0.0377} & \textbf{0.0068} \\
\textbf{GoT}~\cite{fang2025got} & 0.0730 & 0.0287 & 0.0819 & 0.0351 & 0.0564 & 0.0221 & 0.0587 & 0.0219 & 0.0675 & 0.0269  \\
\textbf{ICEdit}~\cite{zhang2025context} & {\ul 0.0422} & {\ul 0.0095} & 0.0411 & 0.0104 & 0.0471 & 0.0131 & {\ul 0.0397} & 0.0083 & {\ul 0.0425} & 0.0103\\
\textbf{Step1X-Edit}~\cite{liu2025step1x} & 0.0462 & 0.0157 & {\ul 0.0392} & 0.0113 & {\ul 0.0413} & 0.0130 & 0.0498 & 0.0155 & 0.0441 &0.0139 \\
\textbf{Gemini}~\cite{team2023gemini} & 0.0657 & 0.0192 & 0.0559 & 0.0159 & 0.0634 & 0.0197 & 0.0531 & 0.0129 & 0.0595 & 0.0169\\
\textbf{Ours (Gemini-CoT)} & 0.0982 & 0.0348 & 0.0718 & 0.0209 & 0.0823 & 0.0292 & 0.0863 & 0.0273 & 0.0846 & 0.0281 \\
\bottomrule
\end{tabular}
\vspace{-1mm}
\end{table*}

\subsection{Vision Consistency Evaluation}
Table~\ref{table3} presents the vision consistency results, measured by L1 and L2 distances (lower is better) in non-edited regions across various instruction complexities. \textbf{OmniGen}~\cite{xiao2024omnigen} exhibits superior consistency, achieving the best average L1 (0.0377) and L2 (0.0068) scores. It consistently minimizes alterations in unchanged areas across all tested scenarios.
Other models like \textbf{ICEdit}~\cite{zhang2025context}, \textbf{Step1X-Edit}~\cite{liu2025step1x}, and \textbf{UltraEdit}~\cite{zhao2024ultraedit} also demonstrate strong preservation capabilities. UltraEdit is particularly effective in L2 distance, ranking second on average.
Conversely, \textbf{Gemini}~\cite{team2023gemini} and \textbf{Ours (Gemini-CoT)}, despite their leading editing performance, show higher L1/L2 scores here, suggesting their extensive edits might impact surrounding regions more. Models like SEED-LLAMA show the least consistency, with significantly higher error scores. This indicates a potential trade-off between aggressive editing and background preservation, where OmniGen currently offers the best balance.

\begin{figure}[t]
    \centering
    \includegraphics[width=1\linewidth]{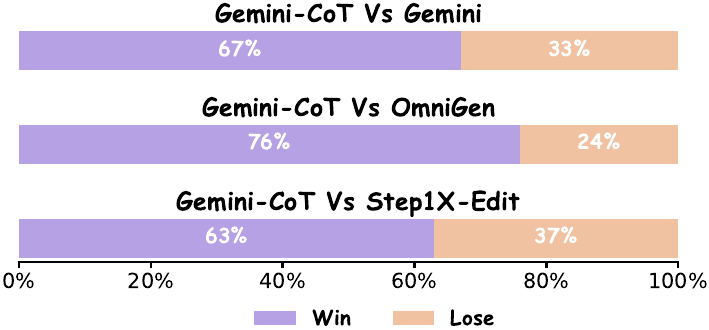}
    \vspace{-7mm}
    \caption{\small Human evaluation between Gemini-CoT and others.}
    \label{fig:human evaluation}
    \vspace{-3mm}
\end{figure}

\subsection{Effect of Vision CoT}
Figure~\ref{fig:case} illustrates CoT reasoning's benefit for complex image editing, exemplified by a multi-step instruction to add and modify oranges. CoT provides a structured thought process (analyzing, deconstructing, planning). With CoT guidance, the Gemini model more accurately executes instructions, adding the whole orange and a green fruit element as intended. Without CoT, it struggles, producing a less faithful result that fails to correctly add a distinct slice or apply the color change. This comparison highlights how CoT enhances the model's ability to understand and execute complex, sequential edits, improving alignment with user intent.

\subsection{Human Evaluation}
To complement automated metrics, we conducted human evaluations assessing the perceptual quality of edits from our Gemini-CoT method against leading models. As shown in Figure~\ref{fig:human evaluation}, evaluators performed pairwise comparisons. Results indicate a strong preference for Gemini-CoT: it was preferred over standard Gemini in 67\% of cases, over OmniGen in 76\%, and over Step1X-Edit in 63\%. These findings highlight CoT's effectiveness in producing edits that are quantitatively superior and better aligned with human perception of quality and instruction adherence.

% \subsection{Comparison With GPT-4o}

\section{Conclusion}
We introduced ComplexBench-Edit, a novel benchmark for evaluating image editing models on complex, multi-instruction tasks, especially chained dependencies. We also presented Gemini-CoT, a training-free method using Chain-of-Thought reasoning, which significantly enhanced the execution of these intricate instructions. Experiments demonstrated Gemini-CoT's superior performance where many existing models struggle, particularly with sequential edits, a finding corroborated by human evaluations.

\bibliographystyle{ACM-Reference-Format}
\bibliography{ref}
\end{document}